\documentclass[10pt,conference]{IEEEtran}
\IEEEoverridecommandlockouts
\usepackage{cite}
\usepackage{amsmath,amssymb,amsfonts}
\usepackage{algorithmic}
\usepackage{graphicx}
\usepackage{textcomp}
\usepackage{xcolor}

\usepackage{graphicx}
\usepackage[hidelinks]{hyperref}

\usepackage{amsmath,amssymb,amsfonts}
\usepackage{algorithmic}
\usepackage{graphicx}
\usepackage{textcomp}
\usepackage{xcolor}
\usepackage{subcaption}

\usepackage{color,xcolor}
\usepackage{epsfig}
\usepackage{graphicx}

\usepackage{adjustbox}
\usepackage{array}
\usepackage{booktabs}
\usepackage{multirow}

\usepackage{amsmath,amsfonts,amsthm,amssymb}
\usepackage{bm}
\usepackage{nicefrac}
\usepackage{microtype}

\usepackage{changepage}
\usepackage{extramarks}
\usepackage{fancyhdr}
\usepackage{lastpage}
\usepackage{setspace}
\usepackage{soul}
\usepackage{xspace}

\usepackage{marvosym}

\def\BibTeX{{\rm B\kern-.05em{\sc i\kern-.025em b}\kern-.08em
    T\kern-.1667em\lower.7ex\hbox{E}\kern-.125emX}}
\begin{document}

\title{Federated Contrastive Learning for Dermatological Disease Diagnosis via On-device Learning
\thanks{This work was supported in part by NSF CNS-2122320.}}

\author{\IEEEauthorblockN{
Yawen Wu\textsuperscript{1},
Dewen Zeng\textsuperscript{2}, 
Zhepeng Wang\textsuperscript{3}, 
Yi Sheng\textsuperscript{3},
Lei Yang\textsuperscript{4},
Alaina J. James\textsuperscript{1,5},
Yiyu Shi\textsuperscript{2},
Jingtong Hu\textsuperscript{1},}

\IEEEauthorblockA{\textsuperscript{1}University of Pittsburgh, PA, USA.
\textsuperscript{2}University of Notre Dame, IN, USA.\\
\textsuperscript{3}George Mason University, VA, USA.
\textsuperscript{4}University of New Mexico, NM, USA.\\
\textsuperscript{5}University of Pittsburgh Medical Center, PA, USA.\\
yawen.wu@pitt.edu, dzeng2@nd.edu, zwang48@gmu.edu, ysheng2@gmu.edu, \\ leiyang@unm.edu, jamesaj@upmc.edu, yshi4@nd.edu, jthu@pitt.edu
}
}

\maketitle

\begin{abstract}
Deep learning models have been deployed in an increasing number of edge and mobile devices to provide healthcare.
These models rely on training with a tremendous amount of labeled data to achieve high accuracy.
However, for medical applications such as dermatological disease diagnosis, the private data collected by mobile dermatology assistants exist on distributed mobile devices of patients, and each device only has a limited amount of data. 
Directly learning from limited data greatly deteriorates the performance of learned models.
Federated learning (FL) can train models by using data distributed on devices while keeping the data local for privacy.
Existing works on FL assume all the data have ground-truth labels.
However, medical data often comes without any accompanying labels since labeling requires expertise and results in prohibitively high labor costs.
The recently developed self-supervised learning approach, contrastive learning (CL), can leverage the unlabeled data to pre-train a model for learning data representations, after which the learned model can be fine-tuned on limited labeled data to perform dermatological disease diagnosis.
However, simply combining CL with FL as federated contrastive learning (FCL) will result in ineffective learning since CL requires diverse data for accurate learning but each device in FL only has limited data diversity.
In this work, we propose an on-device FCL framework for dermatological disease diagnosis with limited labels.
Features are shared among devices in the FCL pre-training process to provide diverse and accurate contrastive information without sharing raw data for privacy. After that, the pre-trained model is fine-tuned with local labeled data independently on each device or collaboratively with supervised federated learning on all devices.
Experiments on dermatological disease datasets show that the proposed framework effectively improves the recall and precision of dermatological disease diagnosis compared with state-of-the-art methods.
\end{abstract}


\section{Introduction}

Skin diseases are a major global health threat to a tremendous amount of people in the world \cite{verma2019classification}.
These diseases not only injure the physical health including the risk of
skin cancer but also can result in psychological problems such as lack of self-confidence and psychological depression due to damaged appearance \cite{ahmad2020discriminative, wu2019studies}.
Deep learning models have shown great promising in skin diseases diagnosis \cite{wu2019studies, velasco2019smartphone,gu2019progressive} and have been widely deployed on mobile devices as mobile dermatology assistants \cite{googleai, firstderm, DermExpert}.
These models are trained on a large amount of data with full labels to achieve a high accuracy \cite{wu2020enabling}. 
When the large-scale datasets for training are not available, the performance of deep learning models will greatly degrade \cite{kairouz2019advances}.
However, the images of skin disease are usually distributed on mobile devices of patients, which are impractical and even illegal to combine in a single location \cite{kairouz2019advances} to form large-scale datasets since data sharing is constrained by the Health Insurance Portability and Accountability Act (HIPAA) \cite{kairouz2019advances}.
For example, skin disease images can be taken by the cameras of mobile devices and stored for a preliminary self-diagnosis \cite{velasco2019smartphone,sun2016benchmark}.
But patients are usually reluctant to share highly private and sensitive images with the data center. 
Without large-scale datasets in a single location, it is not doable to perform centralized training for learning an accurate model.

Federated learning (FL) is a distributed learning framework where many mobile devices collaboratively learn a global prediction model without sharing private data \cite{yang2019federated}. 
By leveraging FL, distributed data on mobile devices 
can be used to train an accurate shared model to diagnose skin diseases while keeping data local.
Existing FL works assume the local data on devices are fully labeled and use supervised learning for local model updates. However, the assumption of fully labeled data is impractical. For instance, the patients may not want to spend time labeling their skin images captured by cameras of mobile phones. Even voluntary patients may not be able to accurately label all their own images due to the lack of expertise. 
Therefore, most of the distributed data on devices will be unlabeled, and the deficiency of labels makes supervised FL unrealistic.

Contrastive learning (CL), a recently developed self-supervised learning approach, can learn effective visual representations on data without using labels \cite{he2020momentum}.
By combining CL with FL as federated contrastive learning (FCL), the conventional supervised learning on local devices can be replaced by CL pre-training without using labels.
After that, the pre-trained model can be used as the initialization to fine-tune for the diagnosis task with limited labels.
In this way, an accurate dermatological disease diagnosis model can be learned by using distributed data with limited labels.

However, simply combining CL into FL cannot achieve optimal performance. 
This is because existing CL approaches \cite{he2020momentum, chen2020simple, caron2020unsupervised} are originally developed for centralized training on large-scale datasets, assuming sufficiently diverse data is available for training. 
More specifically, different from supervised learning with cross-entropy loss, in which each image is used independently from other images, CL relies on diverse data to learn the correlation between different images.
Without large data diversity, the performance of CL performance will greatly degrade, which also result in a low accuracy on the skin disease diagnosis after fine-tuning with labeled data.

To address this challenge, we propose an on-device FCL framework to enable effective FL with limited labels. This framework has two stages. 
The first stage is federated self-supervised pre-training. Feature sharing is proposed to improve the data diversity of local contrastive learning while avoiding raw data sharing. Data features encoded in vectors are shared among devices, such that diverse and accurate contrastive information is provided to each device. 
By leveraging the shared features, representations of higher quality are learned on local devices, which improves the quality of the aggregated model in FL.

The second stage is fine-tuning with limited labeled data.
By using the pre-trained model in the first stage as a good initialization, the second stage learns the task of dermatological disease diagnosis by either fine-tuning independently on each device, or fine-tuning collaboratively on all devices by supervised federated learning with limited labels.

In summary, the main contributions of this paper include:
\begin{itemize}
	\item \textbf{Federated contrastive learning (FCL) framework.} 
	We propose an on-device FCL framework to enable effective learning with limited labels for dermatological disease diagnosis.
	FCL pre-trains the model on distributed unlabeled data to provide a good initialization, followed by fine-tuning with a limited number of labeled data to perform the disease diagnosis.
	\item \textbf{Feature sharing for better local learning.} 
	We propose a feature sharing method to improve the data diversity of local contrastive learning while avoiding raw data sharing for privacy.
	The shared features provide more diverse features to contrast with during local learning on each mobile device for better representations.
	\item \textbf{More accurate diagnosis and better label efficiency.} 
	Experiments on dermatological disease datasets with various skin colors show superior diagnostic accuracy and label efficiency over state-of-the-art techniques.
\end{itemize}

\section{Background and Related Work}

\begin{figure}[htb]
	\centering
	\includegraphics[width=0.8\linewidth]{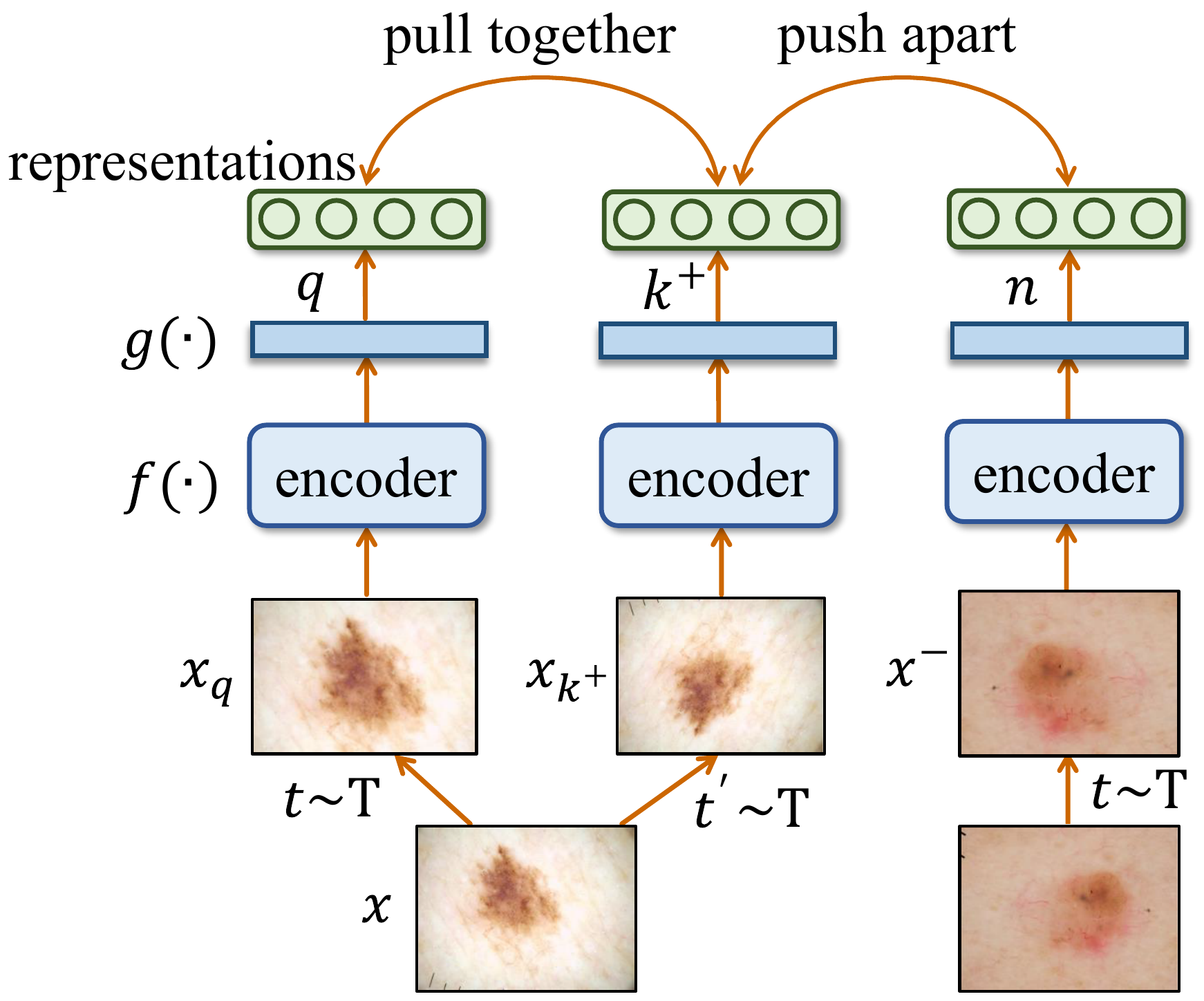}
	\caption{Illustration of contrastive learning. Two separate data augmentations operators are sampled from the same family of augmentations ($t\sim T$ and $t^{\prime} \sim T$) and then applied to one image $x$. The representations of the two transformed versions are pushing close to each other and apart from the representations of other images.}
	\label{fig:contrastive_learning_illustration}
	\vspace{-16pt}
\end{figure}

\begin{figure*}[ht]
	\centering
	\includegraphics[width=1.0\textwidth]{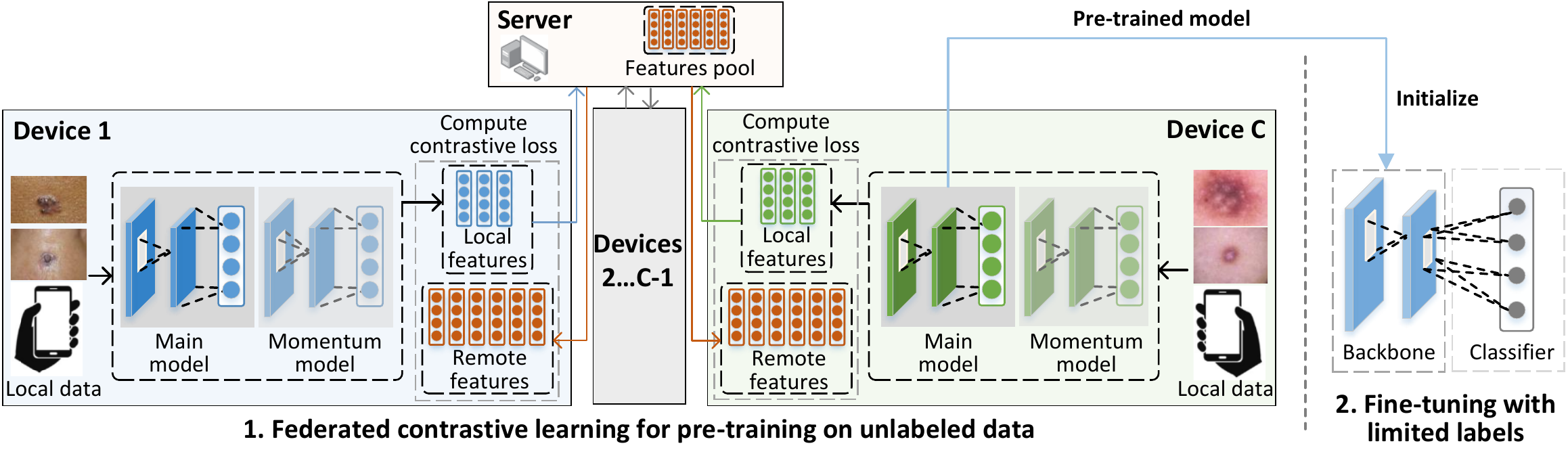}
	\caption{Federated contrastive learning with feature sharing for pre-training the model with unlabeled data, by which good data representations are learned.
    After pre-training, the learned model is used as the initialization for fine-tuning with limited labels for dermatological disease classification. 
	}
	\label{fig:overview}
	\vspace{-12pt}
\end{figure*}

\subsection{Contrastive Learning} 
Contrastive learning (CL) is a powerful self-supervised method to learn visual representations from unlabeled data \cite{tian2019contrastive,misra2020self,tian2020makes,wu2021enabling,chuang2020debiased}.
CL pre-trains a model and provides high generalization performance for downstream tasks such as classification and segmentation \cite{he2020momentum,chen2020simple,chen2020big,zeng2021positional,wu2021federated}.
CL learns representations by performing a proxy task of discriminating the image identities.
In the learning process, CL minimizes a contrastive loss evaluated on pairs of feature vectors extracted from data augmentations (e.g. cropping, rotation, and color distortion) of the image \cite{ho2020contrastive}.
By optimizing the contrastive loss, CL maximizes the agreement of representations between transformations of the same identity and minimizes the agreement between different images \cite{kim2020adversarial}.
As shown in Fig. \ref{fig:contrastive_learning_illustration}, for an unlabeled input image $x$, two random transformations $t \sim T$ and $t^{\prime} \sim T$ are applied to $x$ to produce $x_q$ and $x_{k^{+}}$, both of which are then fed into the model $f$ and projection head $g$ to get representations $q$ and $k^{+}$.
Let $Q$ be a memory bank with $K$ representation vectors stored. By using every feature $n$ in the memory bank $Q$ as negatives, a positive pair $q$ and $k^{+}$ are contrasted with every $n$ by the following contrastive loss function.
\begin{equation}\label{equ:mocoloss}
\ell_{q}=-\log \frac{\exp (q \cdot k^{+} / \tau)}{\exp (q \cdot k^{+} / \tau) + \sum_{n\in Q} \exp (q \cdot n / \tau)}.
\end{equation}
By optimizing the model $f$ to minimize the loss function, effective visual representations can be learned by $f$.

However, existing CL works are developed for centralized training on large-scale datasets consisting of millions \cite{imagenet_cvpr09} or even billions \cite{mahajan2018exploring,he2020momentum} of images.
The large-scale datasets provide sufficiently large data diversity for training.
However, in FL each device only has a limited amount of data with limited diversity. Since CL relies on the contrast with different data to achieve high performance, without large data diversity, the performance of CL will greatly degrade.

\subsection{Federated Learning}
Federated learning (FL) aims to collaboratively learn a global model for distributed devices while keeping data local devices for privacy \cite{mcmahan2017communication}.
In FL, the training data are distributed among devices, and each device has a subset of the training data.
In a typical FL algorithm FedAvg \cite{mcmahan2017communication}, 
learning is performed round-by-round by repeating the local learning and model aggregation process until convergence.
In one round, the server activates a subset of devices and sends them the latest model. Then the activated devices perform learning on local data. 
More specifically, in communication round $t$, the server activates a subset of devices $C^t$ and downloads the latest global model with parameters $\theta^t$ to them.
Each device $c\in C^t$ learns on private dataset $D_c$ by minimizing the local loss 
$\ell_c$ 
to get the updated local parameters $\theta_c^{t+1}$.
The locally updated models are aggregated into the global model by averaging the local parameters  
$\theta^{t+1} \leftarrow \sum_{c\in C^t}\frac{|D_c|}{\sum_{i\in C^t} |D_i|}{\theta_c^{t+1}}$.
This learning process continues until convergence.

The problem with these FL works is that they assume the devices have ground-truth labels for all the data. However, this assumption is not realistic in dermatological disease diagnosis due to the high labeling cost and the requirement of expertise for accurate labeling.
Therefore, to apply FL to dermatological disease diagnosis, one approach to effectively use limited labels to achieve high model accuracy is needed.

\section{Overview of Federated Contrastive Learning}

The overview of the proposed federated contrastive learning (FCL) framework is shown in Fig. \ref{fig:overview}. 
This framework has two stages. 
In the first stage, the model is collaboratively pre-trained by distributed devices on unlabeled data to extract visual representations. 
During learning, there is a server and many devices. The server is used to coordinate the learning process, aggregate the locally updated models and forward the shared features.
The devices perform CL on local data as well as local and remote features to update local models.
In the second stage, the model pre-trained in the first stage is used as the initialization for fine-tuning with limited labels. This can be achieved by existing supervised learning methods independently on each device, or collaboratively by supervised federated learning \cite{mcmahan2017communication}.
In the rest of this paper, we focus on the first stage of FCL on unlabeled data for pre-training.

To achieve better contrastive learning on each device, one can share raw data (i.e. skin images) with other devices to improve the data diversity for local learning. However, since skin images are highly sensitive and private, sharing them will cause serious privacy concerns for patients.
To improve data diversity while keeping raw data local for privacy, we proposed to share features (i.e. encoded vectors).
As shown in Fig. \ref{fig:overview}, FCL is performed round-by-round.
In one round, each device encodes its data as local features and uploads them to the server. Local models are also uploaded for model aggregation.
Then, the server de-identifies the uploaded features and sends these anonymous features to devices as remote features. 
The uploaded models are aggregated by averaging the weights of all models following \cite{mcmahan2017communication}, which serves as the initial model for the next round.
After that, the aggregated model and the remote features are downloaded to devices.
The local models on devices are updated by leveraging both local and remote features on each device, which provides more accurate and diverse contrastive information for computing the local contrastive loss.
Finally, the updated local models and features are uploaded to the server to initialize the next round.
This iterative learning process continues until the model convergence.

\section{Contrastive Learning with Remote and Local Features}

\begin{figure*}[t]
	\centering
	\includegraphics[width=0.85\textwidth]{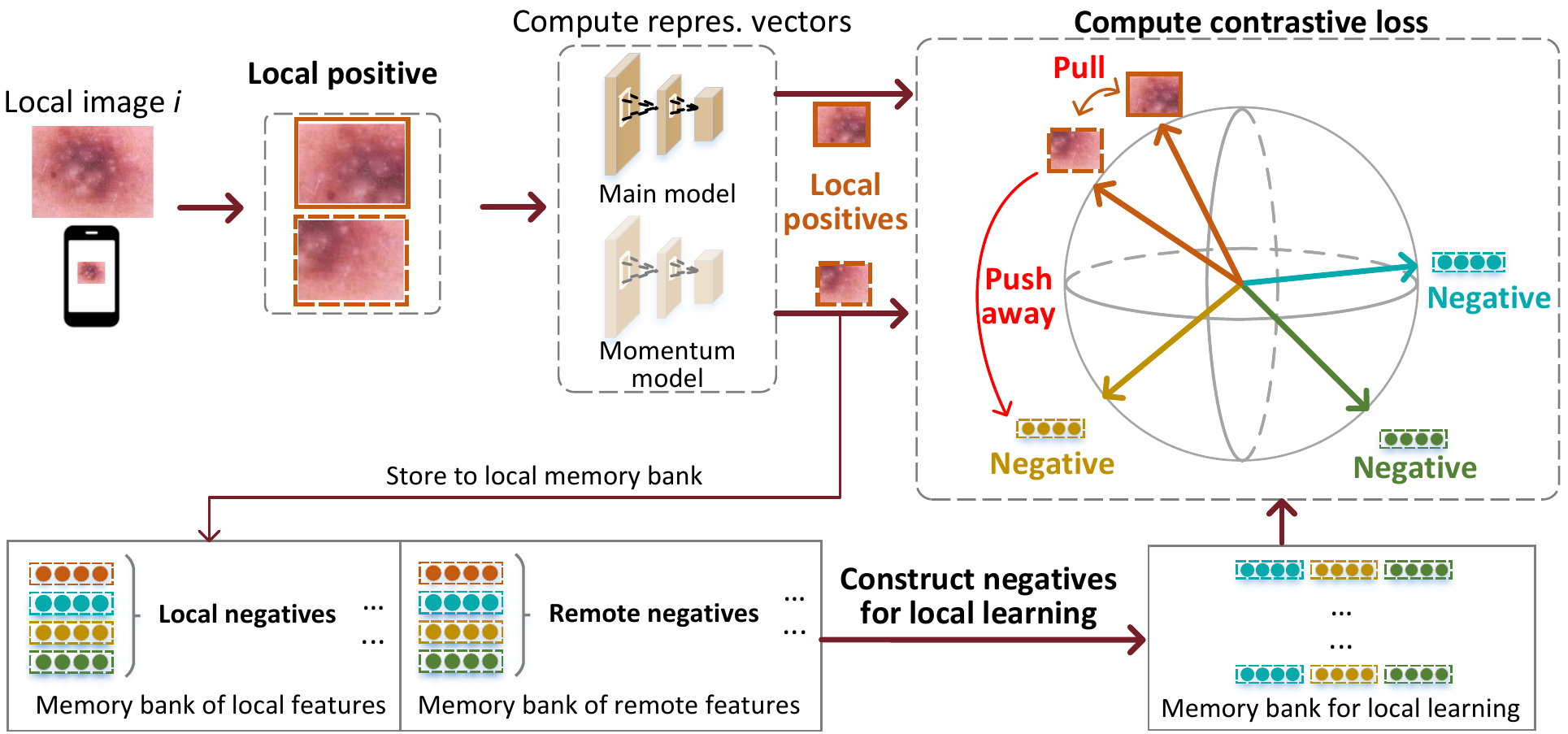}
	\caption{
    Local contrastive learning with shared features on one device in FCL. The remote features from other devices serve as negatives to improve the data diversity for better-learned representations. 
    During learning, the features of local positives are pushed close to each other and apart from the remote negatives.
	}
	\label{fig:feature_exchange}
	\vspace{-12pt}
\end{figure*}

In this section, we will present how to perform contrastive learning on each device with both local and remote features in each round of FCL, by which a good initialization for fine-tuning can be learned.

\subsection{Local Model Architecture}
The local contrastive learning process on a device is shown in Fig. \ref{fig:feature_exchange}.
We use the Momentum Contrast (MoCo) model architecture \cite{he2020momentum} since it uses a separate memory bank to store negative features, which can be filled with both local and remote negatives in the FCL setting.
Other popular contrastive learning models such as SimCLR \cite{chen2020simple} uses features of data in the same mini-batch as negatives, which tightly couples the negative features with the mini-batch data and is not suitable to leverage remote features in FCL.
Different from SimCLR, MoCo decouples the negative features with the mini-batch data by an independent memory bank and can leverage the remote features as negatives for local contrastive learning.

On each device, 
there are two models, the main model and the momentum model, and they have different functions.
The main model is the target model for learning and will be used as the initialization for the fine-tuning in the second stage.
The momentum model is used to generate features for local contrastive learning as local negatives and for sharing as remote features. 
The generated features will be stored in the memory bank of local features.
In the learning process, the main model is updated by the contrastive loss, and the momentum model is updated as the exponential moving average of the main model with a momentum coefficient.

\subsection{Memory Banks of Local and Remote Features}
To perform contrastive learning on each device,
the memory banks of local features and remote features need to be constructed such that they can provide features for computing the local loss. 
Each device has a memory bank of local features and a memory bank of remote features.
Local features are generated by feeding local images to the momentum model to produce the feature vectors. 
The feature vectors are then stored in the memory bank of local features.
On device $i$, let $Q_{l,i}$ be the memory bank of local features with $K$ features stored. $Q_{l,i}$ is used as local negatives and maintained following a first-in-first-out principle. The oldest features in the local memory bank will be dropped when new features are generated.

The remote features are collected from other clients.
At the beginning of each round of FCL, the local features are uploaded to the server and the remote features will be downloaded to fill the memory bank of remote features. More specially, the memory bank of remote features $Q_{r,i}$ on device $i$ is filled as follows.
\begin{equation}\label{equ:remote_neg}
Q_{r,i} = \{Q_{l,c}\ |\ 1 \le c \le |C|, c\ne i \}.
\end{equation}
where $C$ is the set of all devices.

\subsection{Learning with Remote Features}

\noindent
\textbf{Constructing negatives from memory banks for local learning.}
By leveraging memory banks of local features $Q_{l,i}$ and remote features $Q_{r,i}$, the negatives $Q_{\text{CL},i}$ for computing the contrastive loss is constructed as follows. 
For conciseness, we leave out the device index $i$ in $Q_{\text{CL},i}$.

On device $i$,
at the beginning of each round of FCL, $Q_{\text{CL}}$ is initialized as the memory bank of local features $Q_{l,i}$. 
During each mini-batch of local learning, a mini-batch data $x$ of size $B$ is fed into the momentum model to generate the features $q_{l,i,B}$. Then $B$ remote features are sampled uniformly from remote features $Q_{r,i}$ as follows.
\begin{equation}\label{equ:contrastive_negatives}
    q_{r,i,B}=\{ Q_{r,i,j} |\ j \sim \mathcal{U}(|Q_{r,i}|,B) \}.
\end{equation}
where $j \sim \mathcal{U}(|Q_{r,i}|,B)$ means 
$B$ indices are sampled uniformly from the range $[0,|Q_{r,i}|-1]$ and $Q_{r,i,j}$ is the $j$-$th$ feature in $Q_{r,i}$.

After learning a mini-batch, the oldest features in $Q_{\text{CL}}$ will be replaced by the latest features $q_{\text{update}}$, which are constructed as follows.
\begin{equation}\label{equ:q_update}
q_{\text{update}} = \{q_{l,i,B} \cup q_{r,i,B}\}.
\end{equation}

\noindent
\textbf{Removing local negatives for more accurate learning.}
While $Q_{\text{CL}}$ updated by Eq.(\ref{equ:q_update}) contains remote features for improved data diversity, $Q_{\text{CL}}$ also contains local features. 
For better local learning, we propose to completely avoid using local features as negatives during local learning. 
The intuition is that local features can share certain levels of similarity with the data that is being learned because they are from the same patient.
Using local features as negatives can degrade the learned representations since it pushes the representations of the data being learned apart from the local features, which could have been clustered for better representations.

To solve this problem, we eliminate the use of local features during local contrastive learning. At the beginning of each round of FCL, $Q_{\text{CL}}$ is initialized as the memory bank of remote features $Q_{r,i}$ in Eq.(\ref{equ:remote_neg}) instead of $Q_{l,i}$.
The latest features for updating $Q_{\text{CL}}$ are also simplified as follows.
\begin{equation}\label{equ:q_update_no_local_neg}
q_{\text{update}} = \{q_{r,i,B}\}.
\end{equation}

Compared with Eq.(\ref{equ:q_update}), the local negatives are removed in Eq.(\ref{equ:q_update_no_local_neg}).
In this way, better representations can be learned during FCL, which also results in a higher accuracy of the diagnostic model after fine-tuning.

\noindent
\textbf{Loss function.}
On device $i$, by using the constructed $Q_{\text{CL}}$, the feature $q$ of one image being learned is compared with all features in $Q_{\text{CL}}$, and the contrastive loss for $q$ is defined as follows.
\begin{equation}\label{equ:loss_memorybank}
\ell_{q,k^{+},Q_{\text{CL}}} = - \log \frac{\exp (q \cdot k^{+} / \tau)}{\exp (q \cdot k^{+} / \tau) + \sum_{n\in Q_{\text{CL}}} \exp (q \cdot n / \tau)}.
\end{equation}
where the operator $\cdot$ is the dot product between two vectors and $\tau$ is the temperature to control the distribution concentration degree \cite{hinton2015distilling}.
By minimizing the loss, the representations of local data can be effectively learned.

\begin{table*}[!htb]
	\centering
	\caption{Results of \textbf{local fine-tuning} by the proposed approaches and baselines. The model is pre-trained by different approaches without using labels and then fine-tuned with limited labeled data independently on each device. $L$ is the label fraction on each device for fine-tuning. 
	The recall and precision averaged over all devices are reported, and on each device, the recall and precision are averaged over all classes.
	With different label fractions, consistent improvements by the proposed approaches over the baselines are observed.
	}
	\label{tab:exp_local_finetune}
	\resizebox{0.76\linewidth}{!}{
    \begin{tabular}{lcccccccc}
    \toprule
    \multirow{2}{*}{} & \multicolumn{2}{c}{$L$=10\%} & \multicolumn{2}{c}{$L$=20\%} & \multicolumn{2}{c}{$L$=40\%} & \multicolumn{2}{c}{$L$=80\%} \\
    Methods     & Recall & Precision & Recall & Precision & Recall & Precision & Recall & Precision  \\ \midrule
    Random init & 21.56 & 17.35 & 23.13 & 20.79 & 24.88 & 23.69 & 23.92 & 26.06 \\
    Local CL \cite{chaitanya2020contrastive}   & 26.57 & 26.54 & 28.82 & 28.20 & 31.46 & 30.49 & 34.27 & 30.71 \\
    FedRotation \cite{gidaris2018unsupervised} & 23.45 & 25.27 & 25.05 & 25.05 & 29.39 & 28.48 & 32.87 & 28.15 \\
    FedSimCLR \cite{chen2020simple}  & 30.11 & 31.25 & 32.77 & 31.67 & 35.50 & 33.09 & 37.58 & 33.85 \\
    Proposed    & \textbf{30.41} & \textbf{33.54} & \textbf{34.41} & \textbf{32.96} & \textbf{37.03} & \textbf{34.95} & \textbf{39.25} & \textbf{35.02} \\ \bottomrule
    \end{tabular}
    }
\end{table*}

\section{Experiments}

\noindent
\textbf{Datasets.}
The proposed methods are evaluated on four datasets of different skin colors, including the ISIC 2019 challenge dataset \cite{ISIC2019} mainly for white skins, AtlasDerm \cite{AtlasDerm} and Dermnet \cite{Dermnet} mainly for brown skins, and DarkDerm mainly for dark skins collected by us.
Since these four datasets have a different number of diagnostic categories, to form a unified classification task, we use their intersection of five diseases, including basal cell carcinoma (BCC), dermatofibroma (DF), melanoma (MEL), melanocytic nevus (NV), and squamous cell carcinoma (SCC). 
ISIC dataset consists of about 25k dermoscopic images among nine different diagnostic categories, and we use a subset with about 21k images in the above five classes.
AtlasDerm has about 11k images in 560 categories, and we use its subset in the above five classes with 618 images.
Dermnet consists of images in 23 types of dermatology diseases, and we use a subset in the above five classes with 276 images.
DarkDerm is established with 216 images in the above five categories.
In the pre-processing, the images are resized with bi-linear interpolation such that the dimension of the shorter edge is 72 pixels while keeping the original aspect ratio.

\noindent
\textbf{Federated setting.}
We use 10 devices for FL and distribute datasets based on skin colors to simulate different patients.
The ISIC dataset is randomly split into 7 partitions and each partition is distributed to one of the first 7 devices.
The AtlasDerm, Dermnet, and DarkDerm datasets are assigned to one of the following three devices, respectively.
On each device, the assigned dataset is randomly split into a training set and a test set with 60\% and 40\% data, respectively. The test set is not used in any stage of pre-training or fine-tuning.
We use ResNet-18 \cite{he2016deep} as the backbone model, and use a 2-layer MLP projection head to project the representations to 128-dimensional features.

\noindent
\textbf{Evaluation.}
We use the proposed FCL method to pre-train the model by the distributed devices without using labels.
Then the pre-trained model is used as the initialization for fine-tuning with limited labels.
We consider two practical settings for fine-tuning, \textit{local fine-tuning} and \textit{federated fine-tuning}. 
In local fine-tuning, each device independently fine-tunes its model with its limited labeled data after pre-training.
In federated fine-tuning, devices collaboratively fine-tune the pre-trained model with limited labeled data by supervised federated learning.
During fine-tuning, we evaluate with different fractions of labels, where the percentage of labeled data in the training set is $L\in \{10\%,20\%,40\%,80\%\}$ on each device.
Following \cite{wu2019studies}, we use two metrics for evaluation, including the mean recall of each class (i.e. balanced multiclass accuracy used as the primary metric for the ISIC 2019 challenge \cite{ISIC2019}) and the mean precision of each class.
We report the mean recall and mean precision on the test set of all devices.

\noindent
\textbf{Training details.}
The pre-training by the proposed FCL is performed for 100 communication rounds, and FedAvg \cite{mcmahan2017communication} is employed as the model aggregation algorithm on the server in each round.
The ratio of active devices per round is 1.0 and the number of local training epochs before each aggregation is 1.
The batch size is 128 and the initial learning rate is 0.03 with a cosine decay schedule.
In the fine-tuning stage, the model is trained for 20 epochs in local fine-tuning or 100 rounds in federated fine-tuning.
In local fine-tuning, Adam optimizer is used with a batch size 256, a learning rate of 0.0001 with a decay factor of 0.2 at epoch 12 and 16.
In federated fine-tuning, Adam optimizer is used with a batch size 128 and a learning rate of 0.0001.
The training is performed on one Nvidia RTX 2080Ti GPU.

\noindent
\textbf{Baselines.}
We compare the proposed techniques with four baselines for pre-training.
\textit{Random init} uses random model initialization for fine-tuning.
\textit{Local CL} pre-trains the model by contrastive learning independently on each device with unlabeled data.
\textit{Rotation} \cite{gidaris2018unsupervised} is a self-supervised learning approach for pre-training by predicting the rotation angles of images.
\textit{SimCLR} \cite{chen2020simple} is a SOTA contrastive learning based approach.
We combine these two methods with the FL framework FedAvg \cite{mcmahan2017communication} as \textit{FedRotation} and \textit{FedSimCLR}.

\subsection{Local Fine-tuning}

We compare the performance of different methods by local fine-tuning with limited labels. The results are shown in Table \ref{tab:exp_local_finetune}, and both recall and precision are reported.
The proposed methods effectively improve the recall and precision with different fractions of labeled data.
First, with 10\%, 20\%, 40\% and 80\% labeled data on each device for fine-tuning, the proposed approaches outperform the best-performing baseline by 0.30\%, 1.64\%, 1.53\%, 1.67\% for recall, and 2.29\%, 1.29\%, 1.86\%, 1.17\% for precision, respectively.
Second, the proposed approaches effectively use limited labels for fine-tuning. 
With 40\% labels, the proposed approaches achieve 37.03\% recall and 34.95\% precision, which are on par with or even better than the best-performing baseline with $2\times$ labels (37.58\% recall and 33.85\% precision).

\begin{table*}[!htb]
	\centering
	\caption{
	Results of \textbf{federated fine-tuning} by the proposed approaches and baselines.
	The model is pre-trained by different approaches without using labels and then fine-tuned with limited labeled data collaboratively by all devices. 
	$L$ is the label fraction for fine-tuning on each device. 
	The recall and precision averaged over all classes are reported.
	With different label fractions, consistent improvements by the proposed approaches over the baselines are observed.
	}
	\label{tab:exp_federated_finetune}
	\resizebox{0.76\linewidth}{!}{
    \begin{tabular}{lcccccccc}
    \toprule
    \multirow{2}{*}{} & \multicolumn{2}{c}{$L$=10\%} & \multicolumn{2}{c}{$L$=20\%} & \multicolumn{2}{c}{$L$=40\%} & \multicolumn{2}{c}{$L$=80\%} \\
    Methods     & Recall & Precision & Recall & Precision & Recall & Precision & Recall & Precision  \\ \midrule
    Random init & 43.15 & 38.97 & 45.63 & 40.41 & 50.73 & 44.66 & 55.61 & 46.35 \\
    Local CL \cite{chaitanya2020contrastive}   & 43.41 & 39.59 & 45.69 & 41.39 & 50.35 & 45.58 & 55.47 & 47.70 \\
    FedRotation \cite{gidaris2018unsupervised} & 43.18 & 39.57 & 44.36 & 40.27 & 50.69 & 44.10 & 54.27 & 46.70 \\
    FedSimCLR \cite{chen2020simple}  & 45.89 & 41.44 & 48.92 & 43.39 & 54.17 & 46.71 & 58.64 & 48.27 \\
    Proposed    & \textbf{48.03} & \textbf{42.87} & \textbf{51.50} & \textbf{45.71} & \textbf{55.13} & \textbf{48.73} & \textbf{59.23} & \textbf{50.21} \\ \bottomrule
    \end{tabular}
    }
\end{table*}

\subsection{Federated Fine-tuning}

We compare the performance of different methods by federated fine-tuning on all devices with limited labels. The results are shown in Table \ref{tab:exp_federated_finetune}, and both recall and precision are reported.
First, the proposed methods outperform the baselines by a large margin with different fractions of labeled data.
With 10\%, 20\%, 40\% and 80\% labeled data on each device for collaborative fine-tuning, the proposed approaches outperform the best-performing baseline by 2.14\%, 2.58\%, 0.96\%, 0.59\% for recall, and 1.43\%, 2.32\%, 2.02\%, 1.94\% for precision, respectively.
Second, the proposed approaches effectively improve the labeling efficiency. 
For instance, with 10\% labels, the proposed approaches achieve a similar performance as the best-performing baseline with $2\times$ labels (48.03\% vs. 48.92\% for recall and 42.87\% vs. 43.39\% for precision).

\subsection{Ablation Study}

\begin{figure}[!htb]
	\centering
	\includegraphics[width=0.7\columnwidth]{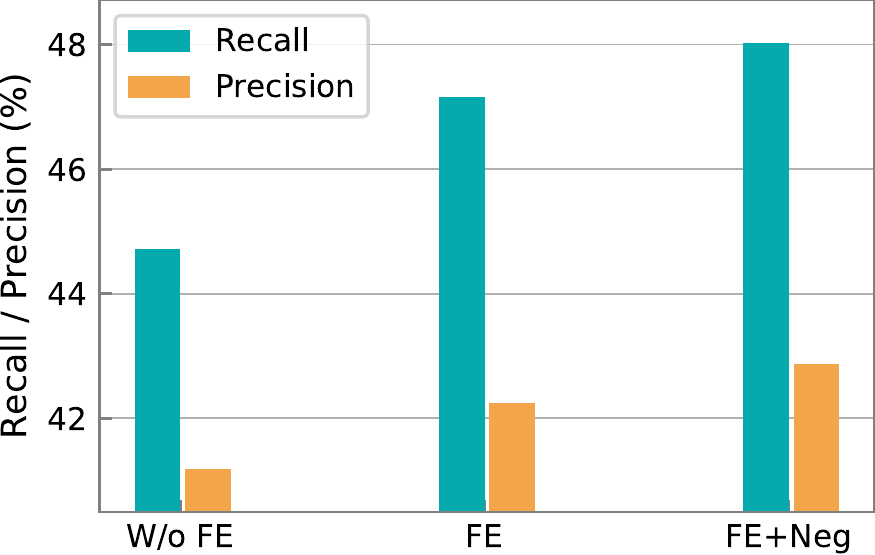}
	\caption{Ablation study. W/o FE is the naive approach without the proposed feature exchange, and FE is the approach with feature exchange enabled. FE+Neg further removes the local negatives. Results by federated fine-tuning with 10\% labeled data are reported. Each of the proposed approaches effectively improves the recall and precision.}
	\label{fig:exp_ablation}
	\vspace{-10pt}
\end{figure}

We perform an ablation study to evaluate the effectiveness of each of the proposed approaches. We evaluate approaches by federated fine-tuning with 10\% labels, and the results are shown in Fig. \ref{fig:exp_ablation}.
For example, without feature exchange, the recall is 44.72\%. By enabling feature exchange, the recall is improved to 47.45\%. By further removing local negatives, the recall is improved to 48.03\%.
This result shows that each of the proposed approaches effectively improves the learned representations and improves the recall and precision for the dermatological disease diagnosis.

\section{Conclusion}
This work aims to enable federated contrastive learning for dermatological disease diagnosis via on-device learning.
In the learning process, devices first collaboratively pre-train a model by using distributed unlabeled data and then fine-tune the model with limited labels.
Feature exchange is proposed to improve the data diversity for better contrastive learning on each device. 
Local negatives are further removed for better clustering of learned representations.
Experimental results on dermatological disease datasets of different skin colors show the effectiveness of the proposed approaches for dermatological disease diagnosis.

\bibliographystyle{IEEEtran}
\bibliography{ref}

\end{document}